\newcommand\SEKIusersusepackages
{
\usepackage{url}
\usepackage{mathpartir}
\usepackage%[pdftex]
{graphicx}
\usepackage{latexsym}
\usepackage{multirow}
\usepackage{pstricks, pst-node, pst-tree,pst-all}
}
\newcommand\SEKIREVIEWSTATUS{1}
\newcommand\SEKIREVIEWER
{\mbox{}
\\J\"org Siekmann
\\FR Informatik, Universit\"at des Saarlandes, D--66123 Saarbr\"ucken, Germany
\\E-mail: {\tt siekmann@ags.uni-sb.de}
\\WWW: \url{http://www-ags.dfki.uni-sb.de}
\\
\\
}
\newcommand\SEKIDOCUMENTCLASS{0}
\newcommand\SEKIYEAR{2008}
\newcommand\SEKINUMBEROFYEAR{01}
\newcommand\SEKITYPE{1}
\newcommand\SEKIPUBLISHEDTYPE{0}
\title{On Requirements for Programming Exercises from an \elearning Perspective}
\author{Carlos Lor\'ia-S\'aenz
        \\ ArtinSoft (\url{www.artinsoft.com})
}
\newcommand{\elearning}{{{e}-Learning\ }}
\newcommand{\elearningnc}{{e}-Learning}
\newcommand{\DAAD}{{DAAD\,}}
\newcommand{\activemath}{{ActiveMath\ }}
\newcommand{\activemathnc}{ActiveMath}
\newcommand{\java}{{\tt Java\ }}

\newcommand{\csharp}{{\tt C\#\ }}

\input seki-deckblatt-2
\usepackage{amssymb}

\mathchardef\Gammaoffont="7000
\mathchardef\Gamma="0100
\mathchardef\Deltaoffont="7001
\mathchardef\Delta="0101
\mathchardef\Thetaoffont="7002
\mathchardef\Theta="0102
\mathchardef\Lambdaoffont="7003
\mathchardef\Lambda="0103
\mathchardef\Xioffont="7004
\mathchardef\Xi="0104
\mathchardef\Pioffont="7005
\mathchardef\Pi="0105
\mathchardef\Sigmaoffont="7006
\mathchardef\Sigma="0106
\mathchardef\Upsilonoffont="7007
\mathchardef\Upsilon="0107
\mathchardef\Phioffont="7008
\mathchardef\Phi="0108
\mathchardef\Psioffont="7009
\mathchardef\Psi="0109
\mathchardef\Omegaoffont="700A
\mathchardef\Omega="010A
\mathchardef\itype="017B

\catcode`\@=11

\gdef\allowhyphens{\penalty\@M \hskip\z@skip}

\gdef\set@low@box#1{\setbox\tw@\hbox{,}\setbox\z@\hbox{#1}\dimen\z@\ht\z@
     \advance\dimen\z@ -\ht\tw@
     \setbox\z@\hbox{\lower\dimen\z@ \box\z@}\ht\z@\ht\tw@ \dp\z@\dp\tw@ }
\gdef\set@low@boxsingle#1{\setbox\tw@\hbox{\rm,}\setbox\z@\hbox{#1}\dimen\z@\ht\z@
     \advance\dimen\z@ -\ht\tw@
     \setbox\z@\hbox{\lower\dimen\z@ \box\z@}\ht\z@\ht\tw@ \dp\z@\dp\tw@ }
\gdef\@glqq{%
\ifhmode\edef\@SF{\spacefactor\the\spacefactor}%
\else\let\@SF\empty
\fi
\CheckFamily\font\fraknomath\ifSameFamily ``\relax
\else\CheckFamily\font\swab\ifSameFamily ``\relax
\else\leavevmode\set@low@box{''}\box\z@\kern-.04em\allowhyphens\@SF\relax
\fi\fi}
\gdef\glqq{\protect\@glqq\kern+.07em}
\gdef\@grqq{%
\ifhmode\edef\@SF{\spacefactor\the\spacefactor}%
\else\let\@SF\empty 
\fi 
\CheckFamily\font\fraknomath\ifSameFamily ''\relax
\else\CheckFamily\font\swab\ifSameFamily ''\relax
\else\kern+.07em``\kern.07em\@SF\relax
\fi\fi}
\gdef\grqq{\protect\@grqq}
\gdef\@glq{{\ifhmode \edef\@SF{\spacefactor\the\spacefactor}\else
     \let\@SF\empty \fi \leavevmode
     \set@low@boxsingle{'\/}\box\z@\kern-.04em\allowhyphens\@SF\relax}}
\gdef\glq{\protect\@glq\kern+.07em}
\gdef\@grq{\ifhmode \edef\@SF{\spacefactor\the\spacefactor}\else
     \let\@SF\empty \fi \kern-.0125em`\kern.07em\@SF\relax}
\gdef\grq{\protect\@grq}

\catcode`\@=12

\makeatletter
\if \@ptsize 0
   \newfont{\scriptscriptscriptgoth}{ygoth scaled 760}
   \newfont{\scriptscriptgoth}{ygoth scaled 833}
   \newfont{\scriptgoth}{ygoth scaled 912}
   \newfont{\gothnomath}{ygoth}
   \newfont{\Goth}{ygoth scaled \magstephalf}
   \newfont{\GOth}{ygoth scaled \magstep1}
   \newfont{\GOTh}{ygoth scaled \magstep2}
   \newfont{\GOTH}{ygoth scaled \magstep3}

   \newfont{\scriptscriptscriptswab}{yswab scaled 760}
   \newfont{\scriptscriptswab}{yswab scaled 833}
   \newfont{\scriptswab}{yswab scaled 912}
   \newfont{\swab}{yswab}
   \newfont{\Swab}{yswab scaled \magstephalf}
   \newfont{\SWab}{yswab scaled \magstep1}
   \newfont{\SWAb}{yswab scaled \magstep2}
   \newfont{\SWAB}{yswab scaled \magstep3}

   \newfont{\scriptscriptscriptfrak}{yfrak scaled 760}
   \newfont{\scriptscriptfrak}{yfrak scaled 833}
   \newfont{\scriptfrak}{yfrak scaled 912}
   \newfont{\fraknomath}{yfrak}
   \newfont{\Frak}{yfrak scaled \magstephalf}
   \newfont{\FRak}{yfrak scaled \magstep1}
   \newfont{\FRAk}{yfrak scaled \magstep2}
   \newfont{\FRAK}{yfrak scaled \magstep3}

   \newfont{\init}{yinit}
   \newfont{\Init}{yinit scaled \magstephalf}
   \newfont{\INit}{yinit scaled \magstep1}
   \newfont{\INIt}{yinit scaled \magstep2}
   \newfont{\INIT}{yinit scaled \magstep3}
\fi
\if \@ptsize 1
   \newfont{\scriptscriptscriptgoth}{ygoth scaled 833}
   \newfont{\scriptscriptgoth}{ygoth scaled 912}
   \newfont{\scriptgoth}{ygoth}
   \newfont{\gothnomath}{ygoth scaled \magstephalf}
   \newfont{\Goth}{ygoth scaled \magstep1}
   \newfont{\GOth}{ygoth scaled \magstep2}
   \newfont{\GOTh}{ygoth scaled \magstep3}
   \newfont{\GOTH}{ygoth scaled \magstep4}

   \newfont{\scriptscriptscriptswab}{yswab scaled 833}
   \newfont{\scriptscriptswab}{yswab scaled 912}
   \newfont{\scriptswab}{yswab}
   \newfont{\swab}{yswab scaled \magstephalf}
   \newfont{\Swab}{yswab scaled \magstep1}
   \newfont{\SWab}{yswab scaled \magstep2}
   \newfont{\SWAb}{yswab scaled \magstep3}
   \newfont{\SWAB}{yswab scaled \magstep4}

   \newfont{\scriptscriptscriptfrak}{yfrak scaled 833}
   \newfont{\scriptscriptfrak}{yfrak scaled 912}
   \newfont{\scriptfrak}{yfrak}
   \newfont{\fraknomath}{yfrak scaled \magstephalf}
   \newfont{\Frak}{yfrak scaled \magstep1}
   \newfont{\FRak}{yfrak scaled \magstep2}
   \newfont{\FRAk}{yfrak scaled \magstep3}
   \newfont{\FRAK}{yfrak scaled \magstep4}

   \newfont{\init}{yinit scaled \magstephalf}
   \newfont{\Init}{yinit scaled \magstep1}
   \newfont{\INit}{yinit scaled \magstep2}
   \newfont{\INIt}{yinit scaled \magstep3}
   \newfont{\INIT}{yinit scaled \magstep4}
\fi
\if \@ptsize 2
   \newfont{\scriptscriptscriptgoth}{ygoth scaled 912}
   \newfont{\scriptscriptgoth}{ygoth}
   \newfont{\scriptgoth}{ygoth scaled \magstephalf}
   \newfont{\gothnomath}{ygoth scaled \magstep1}
   \newfont{\Goth}{ygoth scaled \magstep2}
   \newfont{\GOth}{ygoth scaled \magstep3}
   \newfont{\GOTh}{ygoth scaled \magstep4}
   \newfont{\GOTH}{ygoth scaled \magstep5}

   \newfont{\scriptscriptscriptswab}{yswab scaled 912}
   \newfont{\scriptscriptswab}{yswab}
   \newfont{\scriptswab}{yswab scaled \magstephalf}
   \newfont{\swab}{yswab scaled \magstep1}
   \newfont{\Swab}{yswab scaled \magstep2}
   \newfont{\SWab}{yswab scaled \magstep3}
   \newfont{\SWAb}{yswab scaled \magstep4}
   \newfont{\SWAB}{yswab scaled \magstep5}

   \newfont{\scriptscriptscriptfrak}{yfrak scaled 912}
   \newfont{\scriptscriptfrak}{yfrak}
   \newfont{\scriptfrak}{yfrak scaled \magstephalf}
   \newfont{\fraknomath}{yfrak scaled \magstep1}
   \newfont{\Frak}{yfrak scaled \magstep2}
   \newfont{\FRak}{yfrak scaled \magstep3}
   \newfont{\FRAk}{yfrak scaled \magstep4}
   \newfont{\FRAK}{yfrak scaled \magstep5}

   \newfont{\init}{yinit scaled \magstep1}
   \newfont{\Init}{yinit scaled \magstep2}
   \newfont{\INit}{yinit scaled \magstep3}
   \newfont{\INIt}{yinit scaled \magstep4}
   \newfont{\INIT}{yinit scaled \magstep5}
\fi

\makeatother
\newif\ifSameFamily
\def\CheckFamily#1#2{\GetFamilyName{#1}\ArgOne
        \GetFamilyName{#2}\ArgTwo
        \ifx\ArgOne\ArgTwo\SameFamilytrue\else\SameFamilyfalse\fi}
\def\GetFamilyName#1{\edef\Tempa{#1}\def\Tempb{#1}\ifx\Tempa\Tempb
        \edef\Tempa{\fontname#1}\fi
        \edef\Tempa{\Tempa\space}%
        \expandafter\iGetFamilyName\Tempa\\}
\def\iGetFamilyName#1 #2\\#3{\def#3{#1}}
\def\DefFontName#1#2{{\escapechar-1\expandafter\expandafter\expandafter
        \iDefFontName\expandafter{\csname#2\endcsname}%
        \xdef#1{\expandafter\string\Tempa}}}
\def\iDefFontName{\def\Tempa}

\newcommand\unprotectedae
{\CheckFamily\font\fraknomath\ifSameFamily *a\else
 \CheckFamily\font\swab\ifSameFamily\char'212\else\"a\fi\fi}
\newcommand\unprotectedoe
{\CheckFamily\font\fraknomath\ifSameFamily 
*o\else\CheckFamily\font\swab\ifSameFamily\char'232\else\"o\fi\fi}

\DefFontName\eccclarge{eccc1200}
\DefFontName\eccc{eccc1000}
\DefFontName\ecccsmall{eccc0900}
\DefFontName\ecccfootnotesize{eccc0800}

\renewcommand\ae{\protect\unprotectedae}
\renewcommand\oe{\protect\unprotectedoe}

\newcommand\namefont{}

\newcommand\majorfootroom{\raisebox{-1.9ex}{\rule{0ex}{.5ex}}}

\newcommand\joerg   {J\oe rg}

\newcommand\autexier        {Autexier}
\newcommand\autexiername    {Serge \autexier}

\newcommand\dietrich        {Diet\-rich}
\newcommand\dietrichname    {Dominik \dietrich}

\newcommand\siekmann        {{\namefont Siek\-mann}}
\newcommand\siekmannname    {{\namefont \joerg\ \siekmann}}
\newcommand\wagner          {Wag\-ner}
\newcommand\wagnername      {Marc \wagner}

\newcommand\getittotheright[1]  
{\hfill\mbox{}\penalty 100\mbox{\ \,}\nolinebreak
\hfill\hfill\hfill\nolinebreak\mbox{#1}\ignorespaces}

\newcommand\WWW  {WWW}

\newcommand\CS   {Computer \Sci}

\newcommand\pp   {pp.}
\newcommand\PP[2]{\pp\,\ignorespaces#1--\ignorespaces#2}
\newcommand\Sci  {Sci.}

\newcommand\nthpositioner[2]
{#1\raisebox{0.52ex}{\scriptsize\hspace{0.07em}#2}}

\newcommand\mthpositioner[2]
{\math{#1}\raisebox{0.52ex}{\scriptsize\hspace{0.07em}#2}}
\newcommand\modulointocountzero[2]
{\count1=#1
 \count2=#2
 \count0=\count1
 \divide  \count0 by \count2
 \multiply\count0 by-\count2
 \advance \count0 by \count1}
\newcommand\absolutevalueintocountzero[1]
{\count0=#1
 \ifnum\count0<0\multiply\count0 by -1\fi}
\newcommand\nthstring[1]
{\def\myargone{#1}\ifcat a\myargone th\else\nthstringnochar{#1}\fi}
\newcommand\nthstringnochar[1]
{\absolutevalueintocountzero{#1}%
 \modulointocountzero{\count0}{100} %we need the blank here!
 \ifnum\count0>9\ifnum\count0<20 th\else\stupidnthstring\fi
                                   \else\stupidnthstring\fi}
\newcommand\stupidnthstring
{\modulointocountzero{\count0}{10}
 \ifnum\count0=1 \hskip-0.2em st\else
 \ifnum\count0=2 nd\else
 \ifnum\count0=3 rd\else 
                 th\fi\fi\fi}

\newcommand\writeascents
[1]{\count4=#1
\ifnum\count4<0 
-\multiply\count4 by -1\fi
\modulointocountzero{\count4}{10}
\divide\count4 by 10
\count3=\the\count0
\modulointocountzero{\count4}{10}
\divide\count4 by 10
\the\count4
.\the\count0
\the\count3
}

\newcommand\frenchnthstring[1]
{\def\myargone{#1}\ifcat a\myargone th\else\frenchnthstringnochar{#1}\fi}
\newcommand\frenchnthstringnochar[1]
{\absolutevalueintocountzero{#1}%
 \modulointocountzero{\count0}{100} %we need the blank here!
 \ifnum\count0>9\ifnum\count0<20 th\else\frenchstupidnthstring\fi
                                   \else\frenchstupidnthstring\fi}
\newcommand\frenchstupidnthstring
{\modulointocountzero{\count0}{10}
 \ifnum\count0=1 \hskip-0.2em re\else
 \ifnum\count0=2 me\else
 \ifnum\count0=3 rd\else 
                 th\fi\fi\fi}

\newcommand\CLAM      {{\rm CL\kern-.36em\raise.39ex\hbox{\sc a}\kern-.15emM}}

\newcommand\TEXMACS   {{\sc T\kern-.1667em\lower.5ex\hbox{E}\kern-.125emX\kern-.1em\lower.5ex\hbox{\textsc{m\kern-.05ema\kern-.125emc\kern-.05ems}}}}

\newcommand\ml{{\sc ml}}

\newcommand\academicpress{Academic Press (\elsevier)}

\newcommand\elsevier{Elsevier}

\newcommand\newspaperreference[5]
{\def\nameofjournalpress{#2}#1, #4 #5, #3\if?\nameofjournalpress
 \else, #2\fi}

\newcommand\dateinjournal[1]{}

\newcommand\journalreference[6]
{\def\nameofjournalpress{#2}#1\nolinebreak\hskip.2em%
 \dateinjournal{(#3) }{\mbox{\bf #4}}, \PP{#5}{#6}\if?\nameofjournalpress
 \else, #2\fi}

\newcommand\journalreferenceprintyear[6]
{\def\nameofjournalpress{#2}#1 
 {(#3) }{\bf #4}, \PP{#5}{#6}\if?\nameofjournalpress
 \else, #2\fi}

\newcommand\journalreferenceprintyearaspartofnumber[6]
{\def\nameofjournalpress{#2}#1 
 {#4/#3}, \PP{#5}{#6}\if?\nameofjournalpress
 \else, #2\fi}

\newcommand\jscname
{J. Symbolic Computation}

\newcommand\jscprintyear
{\journalreferenceprintyear{\jscname}\academicpress}

\newcommand\tcsname{Theoretical \CS}
\newcommand\tcsjournal
{\journalreference\tcsname\elsevier}
\newcommand\tcsjournalprintyear
{\journalreferenceprintyear\tcsname\elsevier}

\renewcommand{\ml}{{\tt ML\ }}
\begin{document}
\makecover
% Let your text start here, possibly changing the following a lot.
\maketitle
\begin{abstract}This report accounts for some of the main results obtained by the author during a short research visit at the Saarland University, in Germany.\footnote{Supported by the \DAAD (\url{www.daad.org}), the WG of Prof. Dr. J\"org Siekmann at Saarland University (\url{www-ags.dfki.uni-sb.de}) and ArtinSoft (\url{www.artinsoft.com})} In this work, we deal with the question of modeling programming exercises for novices pointing to an \elearning scenario. Our purpose is to identify  basic requirements, raise some key questions and propose potential answers from a conceptual perspective. We visualize and start delimitating a potential research area, one serving as a basis for further development and more specific work in the future. Presented as a general picture, we hypothetically situate our work in a general context where  \elearning instructional material needs to be adapted to form part of an introductory Computer Science (CS) \elearning course at the {CS1}-level. Meant is a potential course which aims at improving novices skills and knowledge on the essentials of programming by using \elearning based approaches in connection (at least conceptually) with a general \textit{host} framework like \activemathnc\footnote{\url{www.activemath.org}}. Our elaboration covers contextual and, particularly, cognitive elements preparing the terrain for eventual research stages in a derived project, as indicated. Fully aware that this area is huge and multi- and interdisciplinary, we concentrate our main efforts on reasoning mechanisms about exercise complexity that can eventually offer tool support for the task of exercise authoring.  We base our requirements analysis on our own perception of the exercise subsystem provided by \activemath especially within the domain reasoner area.  However, we use other complementary sources, too. We enrich the analysis by bringing to the discussion several relevant contextual elements from the CS1 courses, its definition and implementation. Concerning cognitive models and exercises, we build  upon the principles of Bloom's Taxonomy as a relatively standardized basis and use them as a framework for study and analysis of complexity in basic programming exercises. Our analysis includes requirements for the domain reasoner which are necessary for the exercise analysis. We propose for such a purpose a three-layered conceptual model considering exercise evaluation, programming and metaprogramming.\Keywords{Programming exercises, CS1, \elearningnc, \activemathnc,  Bloom's Taxonomy}\end{abstract}
\section[sec:intro]{Introduction}
We are interested in the general problem of how to take advantage of an \elearning framework (like \cite{Melisetal-ActiveMath-AIEDJ-2001}) as a target for adequately coding programming exercises of the kind we might encounter at a traditional CS1 course level (\cite{acmieee2001}). In such a situation, a core problem to deal with is how \textit{static} learning  programming material (books, exercises, examples, etc.), whose \textit{dynamics} (feedback, diagnosis, guidance, etc.) mainly takes place at classroom or labs or through personal interactions, can be implemented as active \elearning content. In such a way that an important part of the dynamics can become an active component of the corresponding \elearning version. At this work, and from a very particular point of view, we want to contribute to the understanding of some parts of the problem by identifying requirements and proposing solution paths at a concrete level.%
\par In addition, we also consider important to account for reasons and issues at the contextual level because they may have an impact on our understanding of the problem helping us to reflect upon scope, relevance, feasibility and potential benefits of alternative approaches to solution paths. And even tough our work is limited we want to situate it within a practical context and make explicit where it pretends to fit in and where it does not.
\par  Let us begin saying that it is not difficult to become attracted by the potential benefits that \elearning might offer from the point of view of curricula, in general. We may think about things like extensive and massive availability of educational material via the Web and rich electronic media, active course administration, maintenance and application among many other evident advantages which extend the scope of a more traditional classroom oriented model. Hence, given a particularly effective approach to teach programming, it seems quite reasonable to try to delivery it with the help of some \elearning strategy, expecting this way to profit from its benefits at some appreciable degree. To define an adequate strategy for delivering programming is evidently an important and sensitive first stage to take into consideration.%
\par When establishing such an objective at the CS1 level, indirectly  and unavoidably, we come across the well-known debate around what is \textit{the most appropriate way} to teach programming for novices at Computer Science (CS) and Information Technology (IT) related disciplines, where programming is considered in their curricula an essential skill (\cite{acmieee2001}). For answering such a question a lot of conceptions, models and tools have been presented in the past and the  discussion still remains open nowadays (for a small subset where different perspectives and issues are discussed see \cite{acmieee2001,lister2003,jenkins2001,gonzales2006,guzdial2002,mccracken2001,meyer2006, pedroni2006,powers2006}).%
\par Issues in the dispute cover the proper abstract nature of the subject, the definition of minimal programming skills, the programming paradigm, the programming language, the convenience or not of using tools, the approach to motivation and course situation among many others. Such a wide spectrum and space for divergence shows us that the subject CS1 turns out difficult to normalize even though it is technologically, scientifically and socially quite relevant, given the CS/IT impact on many societies.%
\vfill\pagebreak

\par At least, there seems to exist consensus on some of the main issues associated with former and current teaching approaches. For instance, in terms of:
\begin{itemize}
 \item Expected levels of quality from an academic perspective
 \item IT labor market demand and technical requirements 
 \item Student individual motivation for taking a computing discipline as a career (student attraction and retention)
\end{itemize}
These \textbf{dimensions} of learning trigger forces which might naturally compete with each other so the particular teaching approach tends  to  mirror that in one way or another. Therefore, we should not completely ignore them in our analysis.%
\par In fact, contextual considerations of that kind indicate we have enough good reasons for innovation through \elearning techniques at any, at some or even all the mentioned dimensions. We believe that any \elearning approach concerning some key facet of the learning how to program (especially one considered so crucial like programming exercises) should not be designed ignoring some issues related with these forces. Hence, we would consider rather important to try to understand in more detail some of the  key technical issues within the debate (for instance, the question \textit{of the right} programming paradigm and the transition CS1-CS2 under the \textit{functions-first} model, \cite{acmieee2001}). To a  certain extent, we have tried to incorporate them, at least  indirectly, in the background of our analysis.%
\par However, and given the limited scope of this research, we decided to focus on cognitive models of programming in a way that can help us understand more formally why exercise are considered \textit{'hard'} for novices (we use \cite{biggerstaff1993,glaser2000, hundhausen2006,kuittinen2004,lister2006SOLO,lister2003,solloway1986,winslow1996} and \cite{hocgreen1991} as a general and standard reference). We address the question within the frame of \elearningnc.%
\par We do not consider socio-motivational (\cite{gonzales2006,guzdial2002}) or market related concerns (\cite{meyer2006}), and in order to understand a relative small part of the entire problem, we do it in the context of a particular CS1 instructional material just as a case study. Through this input, we try to raise some questions on how such \textit{complexity} is reflected and can be recognized on exercise instances by using a hopefully cognitively adequate and sound set of reasoning criteria.%
\par For such a purpose, we mainly follow approaches based on or quite related to Bloom's Taxonomy (\cite{anderson2001}) because it is a simple yet powerful framework which has been applied in CS1 related questions with apparently quite interesting results (\cite{johnson2007,lister2003a,lister2003,gonzales2006, thompson2008}).%
\par We interprete such experiences found in the literature as a sound basis to understand the complexity of some class of exercises as a part of an \elearning environment. Through our work, we identify interesting questions concerning representation and formal reasoning and contribute with conceptual proposals to deal with them. We consider the \activemath model of exercise (\cite{gouadze2007,zinn2006}) as a point of reference to derive from there our requirements on representation and manipulation languages in terms of a potential implementation.
\vfill\pagebreak

\begin{figure}[h]
\centering
{\small
\begin{verbatim}
Consider the following code fragment:

int[] x1 = {1, 2, 4, 7};
int[] x2 = {1, 2, 5, 7};
int i1 = x1.length-1;
int i2 = x2.length-1;
int count = 0;

while ((i1>0) && (i2>0)){
   if (x1[i1] == x2[i2]) {
      ++count;
      --i1;
      --i2;
   }
   else {
      if (x1[i1] < x2[i2])
         --i2;
      else
         --i1;
   }
}

After the above while loop finishes, 'count' contains what value?
a) 3
b) 2
c) 1
d) 0
\end{verbatim}
}
\caption{Java Version of Leeds Question 2 taken from \cite{lister2006SOLO}}
\label{fig:question2}
\end{figure}

\section[sec:motivation]{An Illustrating Example}
In order to illustrate some of the questions around exercise complexity, we present an example taken from \cite{lister2006SOLO} as shown in Figure~\ref{fig:question2}. The example is referred to as the \java version of Question 2. It was used in the "Leeds"-Group study as explained by the authors in the context of testing program reading and understanding capabilities. Some studies in the past have revealed weaknesses in programmers on relation with these basic skills (\cite{mccracken2001}). The example constitutes a nice case of reasoning about exercise complexity in the way we are interested to formalize.

\par According to the cited work the steps indicated in Figure~\ref{fig:plan_question2} represent a reasonable  solution plan that can be expected to be followed by some responder. The answer a) is the result of following this plan.%

\vfill\pagebreak

\begin{figure}[h]\small
\begin{enumerate}
\item Read through the code
\item Infer that code 
      counts the number of common elements in the two sorted arrays
\item Count manually that number in the two given arrays, namely 3
\item Conclude that option a) is correct
\end{enumerate}
\caption{Plausible Solution Plan to Question in Figure~\ref{fig:question2}}
\label{fig:plan_question2}
\end{figure}

\vfill

\par We notice that counting common elements in sorted arrays is basically the intended function which is implemented by the \java code. As we can judge by seeing the plan of Figure~\ref{fig:plan_question2}, to be able to reach such kind of inference is recognized as an \textit{advanced} skill for programmers at the level of a CS1 course. Apparently, such a skill is here being exercised. 

\par However, the question hides a little trick, a further difficulty which might be easy to overlook: the array indexes \texttt{i1} and \texttt{i2} do not reach the value 0 which in \java is the first index of an array, as known. So the correct answer is actually b). According to the authors 65\% answered correctly, but 21\% chose a). This is an extremely interesting empirical result which is further analyzed in detail by the source using the SOLO Taxonomy, which is worthy to read but is currently beyond our scope.%
\par Let us recapitulate some key ideas from the example. The solution procedure shown above would match the solving-plan that a relative experienced programmer would apply by means of concept/role  assignment techniques in order to recover the intended function (see \cite{biggerstaff1993, kuittinen2004}). In contrast, another less experienced responder would probably use a strategy of a lower cognitive level, essentially by executing the fragment and obtaining the right answer more or less mechanically (\cite{wiedenbeck1999}). Probably, without having \textit{grasped} the intended function, so without needing to understand the code at the abstract level. Thus, the less experienced programmer would get a better performance than the more experienced one,  who fell into the trap of the trick.%
\par The question is which level of complexity is mainly being exercised: recovering the intention or applying the concrete operational semantics of arrays and loops. If recovering was the goal, maybe 65\% of correct answers is not properly reflecting it. If both area were aimed at by exercise, we learn that adding such a small piece of complexity might yield a potentially unexpected result. Tricks like these seen here are known as \textit{distractors}. They turn out useful in exposing some misconceptions or uncovering incomplete learning. They should be consistent with the complexity of the test, however.

\vfill\pagebreak

\par Perhaps the source of such a potential inconsistency, if there is any, could be recognized during the authoring phase of an exercise. Popular cognitive models, like Bloom's Taxonomy, and their interpretations in the programming field, are useful frameworks that can help instructors and teachers to reason about that kind of issue. Hence, in a CS1 or similar \elearning context, it would certainly be  interesting that the computational representation of exercises would allow semiautomated reasoning at that level. The situation we have in mind would be one where some sort of analysis of exercises could be performed in a \textit{static} way during the authoring stage; eventually warning for potential \textit{cognitive inconsistencies}, as in example in Figure~\ref{fig:question2}.%

\par At this point and just briefly, we like to relate our  goal of checking consistency with the approach of the so-called \textit{buggy rules} (\cite{zinn2006}). As we will explain later on, our goal to have a reasoning procedure about exercises can be seen as a technique that would allow to suggest whether something like buggy rules might be required at some point in a given exercise specification. Another related case employing something similar to the notion of buggy rules is \cite{robinet2007}. In both cases, we realize that a key to take into account in our analysis is a domain reasoner on programming exercises. Such a reasoner should allow us to express exercise playing as in \cite{zinn2006} but also additional requirements like reasoning on the exercise itself would then be possible. For instance, in terms of exercise complexity, design and reuse, as we explain in more detail later on.

\vfill

\section[sec:structure]{Structure of the Rest of the Paper}
The rest of this report is  organized as follows. In Section~\ref{sec:why} we present the background on which we base the study. In general, the idea is to review the material looking for requirements according to the scope and goals of the work. We cover definitions and models for CS1 in order to obtain contextually relevant feedback. We also analyze \elearning and the features and facilities of \activemath which are most related with the exercise subsystem. We pay special attention to the idea of a domain reasoner from a conceptual perspective. Then, we address Bloom's Taxonomy and focus on its computational interpretation as a framework for semantic based informal reasoning. We present some cases of study where the taxonomy has been applied for CS1 courses. In Section~\ref{sec:model} we recapitulate the identified requirements and reformulate them as a kind of \textit{agent interaction system}, in a simplified but useful way.  This allows us to derive a conceptualization by means of a three-layered reasoning model. We then show how this architectural model could be used to represent exercise operations and its relation to the requirements of a domain reasoner adequate for programming. Finally, in Section~\ref{sec:metamodel} we illustrate how our model can be used for reasoning about complexity and also we summarize some open problems for further research. And finally, in Section~\ref{sec:conclusions} we summarize and give conclusions pointing to future work.
\vfill\vfill\vfill\vfill\vfill\vfill\pagebreak

\section[sec:why]{Background}\label{sec:why}
We go through the main material we use in this work pointing to its relevance with respect to our goals, especially looking for requirements concerning \elearningnc. Our intention is that this report could serve as a source for further research. In the following, we focus on  programming-in-the-small for learning purposes, additional forms of expressing programs (\cite{powers2006}) are briefly mentioned but really not considered here.
\subsection[sec:acm_ieee_curriculum]{CS1 View under ACM/IEEE}
We take a very brief look at the ACM/IEEE Computing Curricula 2001 (CC2001 for short; \cite{acmieee2001}). It serves here simply as a particular sample model for realizing what a CS1 course might thematically cover and we employ it  as a well-known and widely referred standard where some relevant aspects deserve special attention. Given the scope of this work we just account for two of them in order to identify potential research opportunities for \elearningnc.%
\subsubsection[sec:acmieee_multiprofile]{Multiple Profiles and Expected Performance}
\par We first recall that CC2001 considers 5 disciplines related to computing, namely Computer Science (CS), Computer Engineering (CE), Information Systems (IS), Information Technology (IT) and Software Engineering (SE).%
\par Such a division turns out relevant in our context because CC2001 conceives different profiles with respect to expected performance on programming capabilities, even though CC2001 assumes a \textit{common core} for the first year and values programming as an essential skill.%
\par To see that, let us take a look at Table~\ref{table:acm_ieee_01} which is an small excerpt from Table~3.3 of \cite{acmieee2001} just showing  the portion  named 'Algorithms'.  We notice that columns CS and IS are mostly opposed to each other, according to this model. Moreover, a uniform and interesting difference between CS and SE is clearly expressed, too. Although these are models of expectation on competences from a curricular point of view, they actually may coincide with students expectations and professional aspirations.%
\par In fact, due to organizational aspects in some CS Departments, it happens that students from different computing disciplines attend the same course and thus get exposed to the same programme at the first-year level\footnote{In our personal experience, SE, IS and CS. In student proportions that agree with data \cite{jenkins2001}, in our opinion.}. It is interesting to remember that known studies like \cite{mccracken2001} showed a bimodal\footnote{A big group of \textit{weak} students and a small group of \textit{strong} students.} distribution of the student performance in programming evidencing the lack of normality on assimilation. Thus, first year course implementations may need to cope with a similar situation, which means to pay careful attention to assessment and grading systems, and consequently to exercise models, as a consequence (as done in \cite{lister2003a, lister2003} cognitively sustained on Bloom's Taxonomy).%
\par We identify another opportunity to employ \elearning techniques to help students remain more uniform as a group. Using appropriate exercises in those areas where cognitive differences could negatively influence individual performance. This may avoid, in many cases, the loss of motivation and attrition problems at an early stage. We believe \elearning at CS1 level can be aligned with a \textit{funnel} strategy (\cite{acmieee2001}) which is probably more adequate than a \textit{filter} strategy in case of existing different student orientations toward computing disciplines  or when an orientation is not yet mature enough at this level\footnote{In \cite{gonzales2006}, for instance, more emphasis is paid to psychological and motivational involved aspects by means of active learning likewise under a Bloom's Taxonomy sustained methodology.}.

\begin{table}[htb]\small\centering
\begin{tabular}{ l c c c c c}
\hline \\
\textbf{Performance Capability} & \textbf{CE} & \textbf{CS} & \textbf{IS} & \textbf{IT} & \textbf{SE} \\
\hline \\
Prove theoretical results                 & 3 & 5 & 1 & 0 & 3 \\
Develop solutions to programming problems & 3 & 5 & 1 & 1 & 3 \\
Develop proof of concepts programs        & 3 & 5 & 3 & 1 & 3 \\
Determine if faster solution possible     & 3 & 5 & 1 & 1 & 3 \\
\hline
\end{tabular}	
\caption{Expected Performance Capabilities on Algorithms according to Discipline (Scale: 0=no expectation-5=highest expectation)}
\label{table:acm_ieee_01}
\end{table}
\subsubsection[sec:acm_ieee_paradigm]{Multiple Paradigms}
\par With respect to the theme of programming paradigms CC2001 proposes 6 different course models in the first year depending on the paradigm choice: the so-called procedural-first object-first, functions-first, bread-first, algorithms-first, hardware-first strategies. Among such a variety, a valid choice\footnote{Which is well known in our personal experience using Scheme and Java.} is functions-first, where the standard suggests two courses, CS1 using functional programming (FP) to be followed by CS2 using object-oriented programming (OOP).%
\par Such a transition is interesting, no special indications are given by CC2001. An \elearning bridge FP~$\rightarrow$~OOP appears interesting to implement in a CS1-CS2 context. We are aware that the associated problems have been extensively covered in the past as OOP became a widely-spread programming standard at the software industry. However, some new forces can be able to reactivate it. On the one hand, OOP industrial languages (\java and \csharp mainly) have been adopting known FP features like generics (parameterized polymorphic typing) and lambda-expressions (delegates, continuations) and in some cases interesting academic languages exist where pattern-matching is naturally incorporated (\cite{emir2007}). As a result of these trends we can figure out \elearning techniques helping with these multiparadigmatic trends matter, too.
\subsection[sec:learning]{On \elearningnc\, Elements Notions}
Based on \cite{wikipedia:elearning},  we see that \elearning covers those forms of learning in which instructor, learning material and student might be separated by space or time where the gap between  parts is bridged through the use of a wide spectrum of online technologies, mainly Web-based ones, and a planned and methodic teaching/learning experience. Then, \elearning systems serve to complete or complement more traditional forms of learning in environments which are more appropriated to particular individual conditions, according to \cite{Melisetal-ActiveMath-AIEDJ-2001}.%
\par Precisely, this last work provides us with a rich contextual point of reference where, as a general motivation, we are interested in creating content for a CS1-like course and specifically developing exercises to support feedback and assessment insight an \elearning environment. Consequently, a key issue is that instructional material usually needs to be complemented and implemented in such a way that guidance, coaching and feedback, normally provided by the instructor in person, now need to be part of the learning interaction strategy and the content available in the learning environment. Let us consider this a little bit more in detail with attention to exercises.%
\subsubsection{Elements of Exercise Model}
Source content to be adapted as \elearning material is usually taken from some proven and standard course material like the main book, labs projects and similar sources like exercise banks. Such a task can  involve a lot of detailed and careful work. For example, the methodology of the source material might need to be adapted, as already mentioned, it also requires a formalization of the domain model using some formal language and similar tasks.%
\par In particular, with respect to exercises, we may need to design and implement a dialog-based orchestration using  GUI-gestures and corresponding events and associated feedback, hints responses, etc. Such kind of elements need to be assembled in an adequate way. This is naturally an extremely simplified view, we refer to \url{http://www.examat.org} and to \cite{gouadze2007} for appreciating specific details and available tools that aid the design process by hand.%
\par Object representation in \activemathnc relies on OMDoc which is specifically designed for the mathematical domain (\cite{kohlhase2008}). Currently there is a version for representing programs in the same direction: \cite{kohlhasecodeml2008}. We will be referring to the issue as a requirement, later on.%
\par An exercise is essentially modeled by means of a finite state automaton (FSA), according to \cite{gouadze2007}. States in FSA represent interaction stages during exercise playing and edges refer to event-based transitions associated with possible user answers to GUI-elements, those offered by the corresponding state. An example of such an event occurs when the user selects an option from list representing a multiple choice question (MCQ), which is an interaction element that is available in the design tool.%
\begin{figure}
\small
\begin{verbatim}
Consider the following program:
   val x = 7 + 4
   val y = x*(x-1)
   val z = ~x*(y-2)
a) Which identifiers, constants, operators and keywords occur?
b) Which value is bound to the corresponding identifier?
\end{verbatim}

\caption{Version of Exercise 1.1 in \cite{smolka2008}}
\label{fig:example_ml_01}
\end{figure}
\par As a simple illustration, let us consider the code snippet shown in Figure~\ref{fig:example_ml_01}. It is taken\footnote{The original is written in German, the presented version is our free and slightly changed translation.} from the book \cite{smolka2008} which is currently used in an introductory course to programming at the University of Saarland. The book uses functional programming and Standard \ml language.
\par Assume such an exercise would need to be reexpressed in the exercise model of the \activemath environment. Requirements are clear: Example of Figure~\ref{fig:example_ml_01} tests some previously introduced \ml  basic concepts like 'program', 'value', 'identifier', 'operator', etc. (that is concepts near to the lower syntax level). We also notice that some initial notions are required proper of the \ml evaluation strategy: \footnote{For instance, assuming the notion of a typical read-eval-print loop based interpreter has been introduced.} basic operational semantics like statement sequence (composition, ordering) and definition and binding of identifiers. Those concepts and corresponding relations obviously need to be part of a formal domain  model so that the student interacting with the exercise is able to consult (recall) and reinforcement them. Let us assume such concepts are available.%
\par The exercise implementation is a matter of design: we may decide using a MCQ in each case a) and b) and/or textboxes to gather the input of the required answers, among others interaction and presentation elements. FSA stages need to be designed correspondingly. Thus, for instance, in case b) a state can be related to the expression \texttt{val z = ~x*(y-2)} asking for the value of both \texttt{z} and \texttt{y}; this can done in order to test the misconception of reading a variable \footnote{Some students tend to think after a variable is assigned it might loose its value (\cite{kuittinen2004}).}. As can be easily realized, many choices are possible, even if we are dealing with a relatively small exercise, in appearance. A task like this one needs to be performed and evaluated following some parameters and criteria concerning adequacy, alignment and quality. We are interested in studying some of such criteria but not directly addressing the design of the interaction/presentation level.
\subsubsection[sec:complexity_question]{Static Exercise Complexity}
We have seen that an exercise can be implemented as a kind of interaction system whose design and implementation is actually driven by a problem  as its 'leitmotif'. Thus, before or as a part of such an implementation, but at a more abstract level, we might be interested in asking how can we analyze the inherent complexity of the problem but in terms of a general cognitive model, (i.e. independently of the UI-interaction-model). In other words, our question is:
\begin{quote}
\begin{itemize}
\item[] How can we represent  an exercise so that, as much as possible, it can be \textit{statically} \textit{analyzed} in terms of its complexity? The complexity should be derived from its problem specification and surrounding domain model.
\end{itemize}
\end{quote}
How complex an exercise can be is something that naturally depends, in the end, on the person that will be trying to solve it as well as on some environmental conditions. From that point of view, complexity is relative and clearly a \textit{dynamic} value. In fact, it is a function of the student model component to help approximate and diagnose a personal profile as a result of his/her interaction with the problem. However,  exercises have to be correctly \textit{aligned} with \textit{prototypical} student profiles, in order to be effective (\cite{anderson2001}). So we may want to judge complexity relative to some intentionally targeted static profile, under some ideally static conditions.%
\par For instance and using once again Example~\ref{fig:example_ml_01}, we may intuitively reason and anticipate that question a)  could be considered simpler than b); because, for a prototypical student, identifiers and operators notions relate very well to variables and basic arithmetic operations (up to the use of '{\verb ~ }' in \ml instead of a prefixed minus).  But data-binding and control as well as sequential data-flow can be something new or not yet assimilated by such a student. See the interesting experiments evidencing some misconceptions that novices might initially develop with respect to binding (variable assignment, \cite{kuittinen2004})
\footnote{We personally observed following case: in a  similar context, using an eager language, we asked a student to evaluate something like {\tt x*(x-1)} step by step after a definition of {\tt x}, let's say, {\tt val x = 7+4}. We remember he started replacing {\tt x} by {\tt 7+4}. That is, the first step was like: {\tt (7+4)*((7+4)-1)}. Such kind of harmless 'misconception' (from the programming language perspective) should be recognized early before it gets accumulated and becomes potentially harmful.}.%
\par Thus, because part b) of the exercise aims at the basics of data-binding and data-flow and this task can be considered  cognitively  \textit{harder} than part a), we might take that knowledge into account during the design phase and organize the interactions accordingly, maybe as two independent exercises.%
\par Thus, from a pragmatic perspective, we believe complexity could be relevant in a practical situation where relatively many given exercises might need to be carefully adapted to fit in an \elearning environment.
\subsubsection[sec:dynamic_generation]{Dynamic Generation of Exercises}
So far, we have seen a static perspective of exercise generation as done in \activemathnc. This view corresponds to the (pre)authoring facility when it is performed manually. However, as described in \cite{zinn2006}, \activemath also provides a more powerful functionality to dynamically generate exercises using a domain oriented representation instead of a dialog-oriented one. (In \cite{zinn2006} the domain is symbolic differentiation calculus).%
\par Because such a facility represents a  challenging issue in our particular domain of interest, we come back to this issue later on in this work. At this point we just notice that  specific domain reasoning rules for symbolic manipulation of small programs could be required in order to provide a similar capability as in calculus domain.%
\par Just as a simple example, consider case b) of Example~\ref{fig:example_ml_01}. In this case, and certainly depending on the \textit{intelligence} of the exercise generation procedure, we might need rules for expressing data-flow related operations; just as differentiation rules are used in \cite{zinn2006}.%
\par Finally, we want to mention that another similar approach to the automatic generation of exercises is described in \cite{holohan2006}. They report deriving exercises directly from the domain model,  which in that case is the Databases (DB) and SQL. They use a \textit{random-walk} on the DB-Schema representation, exercises are in this case DB queries.
\subsubsection[sec:operations_other_domains]{Domain Specific Reasoning Rules}
As described in \cite{zinn2006} the exercise module in \activemath allows dynamic the generation of exercises where interaction can be added dynamically based on the diagnosis of student answers to the exercise task. Central to this feature is a domain reasoner (DR) working on the exercise task. The domain reasoner behaves like a (rule-based) expert system that provides intermediate solution steps to the task, in order to represent the solution like a plan, actually in the form of a graph. Nodes represent derived subtasks according to the solution proposed by the DR.  Using that information, the exercise module is able to indicate which specific rules the student can apply or compare his/her answers against the expected ones with respect to the solution graph. In addition, buggy-rules can be also included which serve to encode possible common user misconceptions and errors for the specific task. For instance, consider the chain rule for derivation:
\begin{equation}\label{eq:chainrule}
\frac{d[f(g(x))]}{dx} \Rightarrow_{\{expert,chainrule\}} \frac{d[f(z)]}{dz}(g(x)) \times \frac{d[g(x)]}{dx}
\end{equation}
The structure of this domain specific rule lets us formulate a recursive solution plan for a task matching $\frac{d[f(g(x))]}{dx}$ which includes solving both intermediate tasks $\frac{d[f(z)]}{dz}(g(x))$ (outer layer) and $\frac{d[g(x)]}{dx}$ (inner layer).%
\par For instance, $\frac{d [log(sin(x^3))]}{dx}$ is a matching task that evidently reduces using Rule~(\ref{eq:chainrule}) to $\frac{d[log(z)]}{dz}(sin(x^3))$ and $\frac{d[sin(x^3]}{dx}$ as recursive subtasks.%
\par Additionally, incorrect (buggy) rules can be known and applied by the  expert reasoner to identify potential wrong steps like the following when the inner derivation step $\frac{d[g(x)]}{dx}$ is missing:
\begin{equation}\label{eq:chainrule_buggy}
\frac{d[f(g(x))]}{dx} \Rightarrow_{\{buggy, chainrule, inner\_layer\}} \frac{d[f(z)]}{dz}(g(x))
\end{equation}%
In the example below, such a buggy rule would account for explaining some wrong answer from a student, for example like $\frac{1}{sin(x^3)}$ as an answer for $\frac{d [log(sin(x^3))]}{dx}$.%
\par That kind of rules allows the response module to report a missing inner layer as a possible cause of the wrong or incomplete answer.%
\par We also notice that a DR also needs to make use a at lower-level of other tools for symbolic manipulation like a computer algebra system (CAS). For example, to compare symbolic expressions not only syntactically but also using their  normal forms. In the same direction, a matching operation associated with a rule selection can be more involving, because in case some previous symbolic transformation is required to allow a syntactic matching and rule application. Think of AC-matching for instance.
\par Robinet (et al) reports in \cite{robinet2007} a dynamic model to identify so-called \textit{intermediate metal steps} when solving linear equations and inequations like for instance $2x+9=8+6x$ (the target \elearning system is  \texttt{APLUSIX}). In this case, domain rules include moving monomials from one side of equation to the other and algebraic calculations. Notice that rule selection is don't care non-deterministic. If a student would wrongly response, let us say, with $8x=17$ then the model attempts to find the most probable path (a combinations of movements and calculations) explaining the error. Such a path is then used for tuning a student model and diagnosis (for instance, to detect whether student has more problems with movements or with calculations or both).
\vfill\pagebreak

\section[sec:Bloom]{On Bloom's Taxonomy}
We briefly review Bloom's Taxonomy as a standard framework for analyzing educational objectives and this turns to be useful for answering questions concerning exercise complexity, in particular. We restrict the presentation to the cognitive taxonomy. We will use the revised version as presented in \cite{anderson2001}, which is two-dimensional. We believe that it is adequate in our context where procedural programming skills and  knowledge are both relevant. We cover the material using similar notions as in conceptual modeling and semantic based reasoning (\cite{brachman2004}). The reason is that the taxonomy could be conceived as a procedure for formal reasoning on semantic complexity of exercises.%
\subsection[sec:solo]{Basic Notions}
Expressed in our biased computational and simplified perspective, Bloom's Taxonomy is basically an empirical reasoning framework. It deals with the classification  of statements representing objectives, goals and tasks we want someone to achieve and accomplish. Classification is two-dimensional. In principle, it can be used with statements referring to any domain. It is strongly human-oriented, statements are expressed in natural language  and consequently ambiguity can be a problem, as one might expect \footnote{For instance, see \cite{johnson2007} for a study evidencing discrepancies in categorizing questions in the specific case of CS. Similar findings are reported in \cite{thompson2008}. In both cases, however, it is suggested that discrepancies decrease as assumed knowledge of students get clarified among those persons who are categorizing questions.}. Once educational statements get classified,  expert based recommendations (a kind of expert rules) can be used to associate the classification with recommendations about best practices in terms of organizing learning priorities, instruction delivery, assessment and alignment, roughly speaking.%
\par The classification procedure is described as a matrix (See Table~\ref{table:table_bloom_exmaple} for an illustration). Rows represent the \textit{knowledge dimension} with four categories: \textit{Factual}, \textit{Conceptual}, \textit{Procedural} and \textit{Metacognitive}. Columns represent the \textit{cognitive process dimension} where six categories are considered: \textit{Remember}, \textit{Understand}, \textit{Apply}, \textit{Analyze}, \textit{Evaluate} and \textit{Create}. In both cases, a dimension represents a continuum of complexity. Categories in a dimension identify specific ordered  points over this continuum. For instance,  \textit{Apply} is more complex than \textit{Remember} as a cognitive process.  And \textit{Conceptual} is more complex than \textit{Factual} as knowledge. Over this last dimension, complexity usually contrasts concreteness against abstractness, with \textit{Factual} being more concrete than the others.%
\begin{table}[htb]\tiny\centering
 \begin{tabular}{|l|c| c| c| c| c| c|}
\hline
 & \multicolumn{6}{|c|}{\textbf{Cognitive Process}} \\\cline{2-7}
\multirow{2}{*}{\textbf{Knowledge}}   & Remember & Understand & Apply & Analyze & Evaluate & Create \\
%\hline\hline
&&&&&&\\\hline
Factual &\begin{minipage}{1.5cm}
 List primitive data types in a language
\end{minipage}
 & & & & & \\\hline
Conceptual & & & &\begin{minipage}{1.5cm}
Decompose a structured concept in its parts
\end{minipage} & & \\\hline
Procedural & &\begin{minipage}{1.5cm}How to implement a sort algorithm\end{minipage} & & & &\\\hline
Metacognitive & & & & & \begin{minipage}{1.5cm} Criticize learning programming methodology\end{minipage}
&\\
\hline
\end{tabular}
\caption{Taxonomy Table (includes hypothetical examples)}\label{table:table_bloom_exmaple}
\end{table}

\par In order to classify a statement, the procedure first requires us to \textit{normalize} it by putting it in the form of a pair $\texttt{V}\times\texttt{NP}$ for some verb \texttt{V} and noun-phrase \texttt{NP}. Using so-called \textit{clues}, which are association rules, $\texttt{V}\times\texttt{NP}$  can be classified as $\texttt{P}\times\texttt{K}$ for some cognitive process \texttt{P} (mapping \texttt{V}) and knowledge category \texttt{K} (mapping \texttt{NP}). The expression $\texttt{P}\times\texttt{K}$ actually denotes a matrix cell, useful teaching hints and practical strategies, among others, are associated with it. Gathering them is main the purpose of the classification, actually.
\par To show a  simple artificial example, the statement '\textit{To be able to distinguish between an interpreter and a compiler}' could be normalized as {$\texttt{distinguish}\times\texttt{translator}$}, where \texttt{distinguish} is the main verb and {$\texttt{interpreter} \sqcap \texttt{compiler}$} (a composed) NP. Let us assume, it is a subconcept  of concept \texttt{translator}, with respect to some CS knowledge-based domain. According to a well known clue of the framework, \textit{distinguish} is an action whose closest related cognitive process is \textit{Analyze}.%
\par By means of this association,  our original sentence can be expressed in the framework's vocabulary as {$\texttt{Analyze}\times\texttt{ConceptualKnowledge}$}. In terms of complexity, we might infer that for accomplishing such an objective more than just memorizing a plain list of facts is involved: because knowledge contemplates several concepts and their relationships. In addition, analytical thinking skills are also required which are more complex than remembering. Hence, a particular technique to help learning of that kind of knowledge and skill should be used, consequently.%
\par We have used freely a simple symbolic notation in this example just to suggest that a formalization of the procedure is, at least in principle, quite possible and technically interesting, especially when we are trying to find requirements for a tool. So, we  bring to attention that the semantic based inference procedure could be automated to some extent to support the methodology. But naturally the essential question remains:
\begin{quote}
\begin{itemize}How can CS1 exercise \texttt{concepts} be adequately aligned with to Bloom's Taxonomy of verbs, nouns and action rules? What are the right concepts?
\end{itemize}
\end{quote}%
\par Such a question, and posed this way, is too  abstract, indeed. Hence, we want to consider and propose some ideas that we hope, serves to refine it and decompose it into more concrete questions and related issues as well as to identify paths to cope with them.%
\par We summarize the general principles fundamenting Bloom's Taxonomy, expressed according to our particular understanding and objectives:
\begin{enumerate}
 \item Dimensions of learning objectives are knowledge and  processes. They are domain-independent.
 \item Those dimensions can be (linearly) arranged according to categories representing increasingly levels of cognitive complexity 
 \item In order to \textit{reach} one particular category, the inferior category must have been \textit{reached} first
 \item Domain-specific cognitive dimensions can always be semantically related with domain-independent cognitive dimensions. Thus, domain-specific learning objectives satisfy the same principles of cognitive complexity.
\end{enumerate}%
\vfill\pagebreak

\subsubsection[sec:non_standard_bloom_reasoning]{The Need for Non Standard Reasoning}
\par As we already mentioned, sometimes it turns out difficult to make non ambiguous decisions on some particular cases in absence of specific knowledge about the learner as mentioned in the example in \cite{thompson2008}. The question makes sense in a dynamic situation, like the following question implies: how to dynamically judge the complexity of an exercise, statically assumed by its author to assess a particular classification, let us say, {$\texttt{P}\times \texttt{K}$}, if the applier (the student) does not have yet reached level $\texttt{K}$ at the knowledge dimension. What can we conclude in such a case?%
\par Such a kind of question is evidently relevant in the \elearning context, for instance, preparing an appropriated response for an exercise interaction, as we had mentioned before. For such a case, we might hypothesize some explicit \textit{rules} to be used and added to Bloom's framework, which we call  non-standard. For instance, like the following claim:
\begin{quote}
\begin{itemize}
If student exhibits knowledge on a level less than $\texttt{K}$ and the exercise statically assumes {$\texttt{P}\times \texttt{K}$} then the dynamic complexity becomes {$\texttt{Create}\times \texttt{K}$}. In other words, the exercise requires students to create knowledge of level \texttt{K}  in order to perform \texttt{P}.
\end{itemize}
\end{quote}%
\par It is interesting that such a rule can be geometrically interpreted as moving the complexity to the last cell of the row  \texttt{K}, which is the one with the highest skill requirements.%
\par Evidently, it might be debatable whether such a kind of reasoning is empirically valid or not. The point we want to make is that a Bloom's based methodology would need to contemplate similar forms of reasoning when used in a dynamic way as we may have modeling interactions between exercises and students. We identify this as an important open question.
\subsection[sec:bloom_cs1]{Bloom's Taxonomy and CS1}
\par As a case of study, we want to review the way \cite{lister2003, lister2003a} uses Bloom's Taxonomy in CS1. In this case, it is an object-first course using \java and where no \elearning is directly considered. Basically, the main purpose is to define more realistically the course objectives and from there to implement a more aligned grading system\footnote{The original version of the taxonomy is used and as a consequence only the cognitive process dimension}. The author explicitly avoids performing a detailed explicit one-by-one mapping of his statements to the categories of the taxonomy. Instead, he proposes  three general complexity categories which are considered adequate to his CS1 vision and empirical considerations. The proposal is shown in Table~\ref{table:lister_cs1_categories}\footnote{Identifying Bloom's original categories with the verbs used in the revised version one-by-one}.
\begin{table}[hbt]\centering
\begin{tabular}{l l l }
\hline \\
 \textbf{Process Level} & \textbf{Bloom's Levels} \\
\hline \\
 Reading and Understanding    & \textit{Remember} and \textit{Understanding} \\
 Writing small code fragments & \textit{Apply} and \textit{Analyze} \\
 Writing non-trivial programs & \textit{Evaluate} and \textit{Create}\\
\hline
\end{tabular}
\caption{CS1 Cognitive Process Model according to \cite{lister2003}}
\label{table:lister_cs1_categories}
\end{table}
The objectives of CS1 under this model are defined to address the first two categories, only. Thus, we realize that this model pays special attention to develop reading and understanding skills. In fact, that is sufficient for passing it, according to the grading model. The second category is meant as writing code in a well-defined context.%
\par Consequently with this decision, we want to understand it as providing well-defined specifications referring to a domain fully understood by the learner. This should include the appropriate class structure, the signatures and finally the methods should be kept small. So that focus is set as much as possible on expressing well-known concepts at the local (method) level. We want to add, following \cite{glaser2000} among others, that \textit{templates} should be initially used so that the process \textit{Apply} can be incrementally evolved into \textit{Analyze}.%
\par Focusing on reading and understanding offers some benefits by the elaboration of exercises because questions forces the student to solve problems without being able to run code on a machine. The author strongly exploits this by using Multiple Choice Questions (MCQ) as a testing technique. For instance, example in Figure~\ref{fig:example_lister_cs1} in \cite{lister2001} exercises on iterations over arrays.%
\par The exercise design intentionally contains distractors to test misconceptions, as the author explains. In contrast with the exercise shown in Figure~\ref{fig:question2} the intended function is specified, however. Interesting questions remain how to adapt such a kind of exercise in a interactive \elearning environment according to our requirements. We will discuss this, in the second part of this work.%
\begin{figure}[htb]\centering
\small
\begin{verbatim}
public static int maxPos(int[] y, int first, int last){
/* Returns the position of the maximun element in the
*  subsection of the array "y", starting at position "first"
* ending at position "last".
*/
  int bestSoFar = first;
*** missing code goes here ***
}
Which of the following is the missing code from "maxPos"?
(a) for(int i=last;i>first;i--)
       if(y[i] < y[bestSoFar])
         bestSoFar=i;
(b) for(int i=first+1;i<=last;i--)
       if(y[i] < y[bestSoFar])
         bestSoFar=i;
(c) for(int i=last;i>first;i--)
       if(y[i] < bestSoFar)
         bestSoFar=i;
(d) for(int i=last;i>first;i--)
       if(y[i] > y[bestSoFar])
         bestSoFar=i;
(e) for(int i=first+1;i<=last;i--)
       if(y[i] > y[bestSoFar])
         bestSoFar=i;
\end{verbatim}
\label{fig:example_lister_cs1}
\caption{MCQ Exercise on Reading and Understanding in \cite{lister2001}}
\end{figure}
\par It is worth mentioning that the presented CS1 model does not directly match the knowledge dimension of the taxonomy, which poses an interesting question, indeed. Actually, we have not found any CS1 approach dealing with this dimension, directly.%
\par As an attempt to elaborate some ideas for this issue, we briefly consider the learning model used in \cite{barnes2003}. It serves us as a contrast because the CS1 model is different to\cite{lister2003, lister2003a}, and not explictly aligned with Bloom.%
\par Instead, the course model is an implementation of the so-called \textit{early-bird} pattern (\cite{bergin2001}). We may interprete it in terms of driving the course by modeling not directly by language function, which in a SE/OOP context appears to be natural. In fact, authors explicitly aim at SE as a student target.%
\par From the point of view of the Bloom's Taxonomy, we claim, the course is driven by the knowledge dimension, the process dimension is subordinated, in such a sense. Moreover, the knowledge dimension is covered iteratively (like a spiral model), so the course consists of several cycles covering a core set of new topics added on the top or extending previously covered material. Using models and \texttt{Bluej} apparently help to deal with programming language related details, especially at the beginning. In any case, \java plays a central role in this course.%
\par By taking a look at the material in \cite{barnes2003}, we have identified a categorization as shown in Table~\ref{table:proposed_cs1_kcategories}. 
\begin{table}[hbt]\centering
\begin{tabular}{l l p{4cm}}
\hline\\
\textbf{Knowledge} & \textbf{Bloom's Level} & \textbf{Example (Student ...)}\\
\hline\\
 Behavioral& \textit{Factual and Conceptual} & Understands a model of geometric figures with different attributes \\
 Implementation & \textit{Procedural} & Understands  overloading of constructors in \java \\
 Enhancement & \textit{Metacognitive} & Recognize redundancy as a maintenance problem\\
\hline
\end{tabular}
\caption{CS1 Proposed Knowledge Model derived from \cite{barnes2003}}
\label{table:proposed_cs1_kcategories}
\end{table}
We have provided some simple examples as clues as seen in the table. It seems to us, under a model-based approach the \textit{Factual} and \textit{Conceptual} knowledge levels naturally merge building a uniform category. Every cycle starts with a specification of a conceptual model which the student has to understand and \textit{play} with before learn how to implement it (procedurally)\footnote{The tool \texttt{BlueJ} supports consistently such \textit{playing} task.}. Initially the specification is UML-like, later on can be more informal allowing learning of basic design principles as well. In terms of exercising, we find difficult to describe interactions because of the graphical nature of specifications, we are more interested in textually expressed code. However, \cite{barnes2003} provides a lot of exercises at the \texttt{Reading and Understanding} level.%
\par Considering the cognitive level of {\texttt{Writing small code fragments}}, we want to present an exercise which is (mini)pattern-oriented and likewise suitable for exercising refactoring. It also requires exercising some basic understanding. It is shown in Figure~\ref{fig:barnes_refactoring}\footnote{It includes minor changes.}. As we did before, we raise the question about the representation in terms of \elearningnc. Although, the example is quite simple it would demand symbolic generative manipulation of source code if exercises have to be automatically generated could be reused. For instance, a method like the following: 
\begin{center}
\begin{verbatim}
      public int getX(){return x;}
\end{verbatim}
\end{center}
But this hast to be performed independently of the name \texttt{x} and its type \texttt{int}.
\begin{figure}[htb]\centering
\begin{verbatim}
     Add setter and getter methods for the field x in the 
     following class declaration:
     public class A{
         private int x;
     }
\end{verbatim}
\caption{Exercise on Small Code Generation based on \cite{barnes2003}}
\label{fig:barnes_refactoring}
\end{figure}
\par Finally, it is important to account for the model described in \cite{felleisen2003, felleisen1998}. Such a project would pioneer the teaching of Scheme to novices and an adaption of an environment to take beginners learning needs into consideration. As course model, and from a Bloom's based perspective, we consider it shares a lot with \cite{barnes2003} in goals, pattern (recipe) based methodology (very related to \cite{glaser2000}) and particularly the central role a tools plays on incremental learning. A special interesting feature is the existence of different programming languages according to student level. We identify this a requirement in the \elearning context, certainly.
\vfill\pagebreak

\section[sec:model]{A Conceptual Model for Exercises}\label{sec:model}
In the first part of this work we have elaborated what we may call the big picture with respect to requirements on programming exercises. We did it on the basis of several different perspectives. Cognitive elements, case studies, course models and an \elearning environment were considered. In this second part we want to conceptualize and refine those requirements bringing them to a more concrete level. By this means, we want to figure out how they can be further developed pointing to a design and potential implementation, in a continuation of this work.%
\par By now, our goal is to develop a proposal containing general answers to the questions we have raised during the analysis. For such a purpose we present a general framework for reasoning on exercises. Although, our proposal certainly stays at the conceptual and architectural level we are convinced that the main ideas are reasonably comprehensible and straightforward such that the following steps  of a design are easy to appreciate. However, an understanding of the invented requirements is surely necessary for that purpose.%
\subsection[sec:scenarios]{General Scenarios and Assumptions}
We review the requirements and issues we have considered so far. We mostly restrict ourselves to programming exercises and related cognitive elements. Further, we present a strong simplification of scenarios we are modeling, in such a way that we can focus on more specific issues we would like to study and validate. By doing so, we will rephrase the situation in a vocabulary that might not  correspond to the correct one in a strictly technical sense. However, we believe the simplification is fair enough with respect to the sources in this context.%
\begin{enumerate}
 \item We situated the problem of exercises as a part of two scenarios where several actors (we call them agents) interact. Scenarios are agent interaction systems. We have:
\begin{itemize}
 \item The exercise authoring scenario 
\item  The exercise playing scenario
\item  The learning environment.
\end{itemize}
 \item As the main kind of agents we have
\begin{itemize}
 \item An exercise
 \item An author
 \item A tutor
\item A student
\end{itemize}
\item Agents operate on environments that may contain other agents responsible for achieving subsidiary tasks like communication and knowledge gathering. 
\item Initially, exercises can be statically created by an author agent in an authoring environment. The also can be evolve dynamically or be generated automatically as a result of interactions in environments.
\item An exercise contains a question on a (knowledge based) subject aiming at a prototypical student. It also contains a plan to answer the question and a knowledge basis for supporting the question and its answering process. Plans can be composed of alternatives subplans, so we can assume that just one plan is necessary.
\item Student and exercise interact in a playing scenario. The student is committed to solve the exercise. That means, at the end, to accomplish the goals of the plan presented by the exercise. A student also has knowledge (beliefs) and behaves accordingly to its skills and preferences during the playing. The question contained in an exercise can be considered a query to the student beliefs.
\item The question represents a static level of cognitive complexity according to the author. The exercise should be designed to match the complexity and respond appropriately in case that the matching is not occurring as expected. 
\item As a match could fail, the exercise can contain in its plan a course of actions to help the student to accomplish the plan goals. It can choose to recommend alternative paths or submit the student and a diagnosis of its performance to a tutor in a learning environment where new or remedial  interactions can take place.
\end{enumerate}
\subsection[sec:requirements_specific]{Questions and Plans}
The scenarios previously described are indeed quite general, programming is not playing a particular role, yet.  They are also complex, we are not able to further refine them in this work. Therefore, we will concentrate  on the specifics of questions, plans and complexity. In the following, we may sometimes abuse the terminology and identify exercise with questions, even though we are assuming that questions are just components of exercises, in our conceptualization.%
\par Our next step is to cover issues on exercises and plans which essentially means to show a proposal for their representation. Complexity will be handled as a result of modeling decisions concerning the two formers.%
\par Let us now summarize some additional important and more fine requirements:
\begin{enumerate}
 \item Questions in exercises can be expressed in different languages. Some of them are programming languages (PL) some are natural languages (NL) or formal languages (FL) or of other kinds. Combinations of them are possible. We call them specification languages (SL).
\item Some versions of the same kind of exercises could be used in different CS1 course models and PLs. That means, some exercises can be reused and be reexpressed in different languages.
\item Some questions might require executing program fragments, so compilers and interpreters need to be integrated (as CASs are in \activemathnc). We assume learning is related to small programs written in a programming language but without specifying it explicitly. We want the model can be consider the language as a parameter, consequently.
\item Plans are actually expressed as (executable) metaprograms they may refer to expressions occurring in and operate on questions, as well. Therefore, plans require also a corresponding (metaprogramming)language. And once again, reuse can be required at the planning level. 
\item A engine is required for \textit{executing} plans. We consider it actually as a generalization of the concept of domain reasoning, we will name the domain reasoner machine, or simply domain reasoner (DR). Because the PL is a parameter we will denote it by DR\texttt{<}L\texttt{>}, where L is the specification language SL (which usually contains a programming language PL).
\end{enumerate}
\subsection[sec:requirements_complexity]{Assumptions on Complexity}
Our approach to deal with complexity will be based on the following assumption, which appears to be plausible:
 The complexity of a question corresponds to the complexity of the associated plan to answer it in an exercise. Question and plan are the design product of an author. Such a complexity is a function of the cognitive complexity of the operations used in the plan, which are the operations of the machine DR\texttt{<}L\texttt{>} used to execute plans. By mapping those operations to the Bloom's Taxonomy we would be able to assign a complexity to the question. Our goal in the following will be to suggest and illustrate which kind of reasoning could be involved behind those operations that we have in mind for DR\texttt{<}L\texttt{>}.
\vfill\pagebreak

\begin{figure}[h]
\centering
\includegraphics{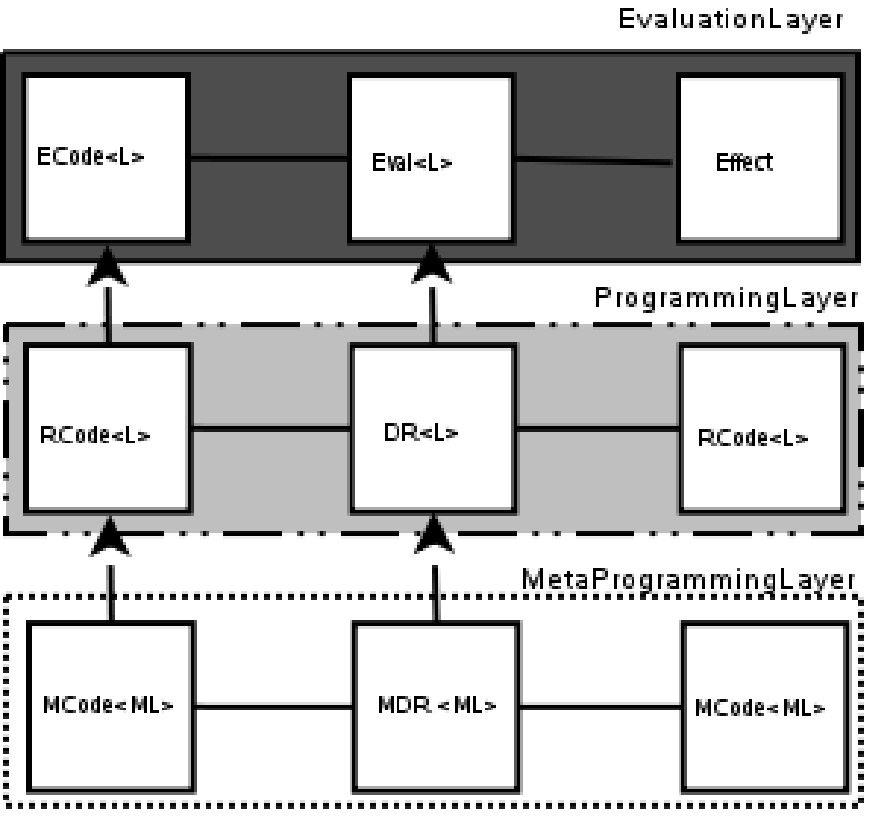}
\vspace*{-1.5em}\caption{Reasoning Architecture}\label{fig:architecture}
\vspace*{1.5em}
\end{figure}

\section[sec:metamodel]{A Layered Reasoning Model}\label{sec:metamodel}
We  are now able to introduce our proposal for a representation of our conceptual model that we will describe  from an architectural point of view. The main idea is shown in Figure~\ref{fig:architecture}. Our representation model is three-layered. We use a metamodeling approach to cope with the requirements of genericity and multi-language capabilities, as facilities we already have identified and indicated, previously. 
\par At the upper level, we have the evaluation layer, where the framework allows adding tools like compilers and interpreters or proxys for them, such that code fragments of the programming language (L) can be evaluated (as CASs systems do in \activemathnc). Expressions in L are reduced to some form of \texttt{effect} object that can be observed by the client of the layer.%
\par Exercises are expressed at the middle layer (called programming layer) where we may have operations working on them which are, intuitively speaking, the plan playing operations. We also will have operations for generating code targeting the evaluation level.%
\par Finally, at the lower level, we have the metaprogramming layer. Here, we  similarly have operations to create metaplans and by this mean  we are able to support plan reuse. A metalanguage ML is used to express plans, code, concepts, which together are called reasoning code (denoted RCode\texttt{<}L\texttt{>}) and metaplans (denoted by MetaCode\texttt{<}ML\texttt{>}), too. We call evaluation code (RCode\texttt{<}L\texttt{>}) the representation of the programming language L.%
\par Based on a metaplan we may generate plan instances that correspond to different exercises of the same kind. Some of them may require  evaluation at the evaluation layer, naturally. We also write Eval\texttt{<}L\texttt{>}, DR\texttt{<}L\texttt{>} and MDR\texttt{<}L\texttt{>} to denote the evaluation/reasoning engines at the corresponding layer. We implicitly intend that they get implemented as rule-based engines, at least at the meta and programming level.
\par This model offers us an additional advantage beyond modularity and genericity features. It give us an interesting metaphor in terms of cognitive models which are compatible with the Bloom's Taxonomy, as we have elaborated in the first part of this work. Thus, conceptually speaking, we may figure out that some student playing an exercise behaves like a kind of DR\texttt{<}L\texttt{>} instance. He/She has to \textit{reason} on code and concepts and relate them with similar/equivalent code and concepts. Maybe code needs to be written and new concepts are introduced during this reasoning process by the student. The tasks involved are exactly the operations the plan associated with the exercise is demanding from him/her to perform.%

\par By following the intended plan, the student can be \textit{forced} during the playing to \textit{go up}, in case the exercise demands executing a program code in L. Or to \textit{go down} and perform an abstraction in order to match, instantiate or create a programming pattern, which require higher-order cognitive capabilities, as we know from Bloom's model. Form this perspective, we believe our abstraction will be reflecting such dynamics in a plausible way.%

\vspace*{3em}

\begin{figure}[h]\centering
\small
\framebox{
\begin{minipage}{10cm}
\tt
What~output~is~produced~by~the~following~Java~code:
\\for(int~i=0;i<=3;i+=2) 
\\~~~System.out.print(i+"~");
\\a)~0~1~2
\\b)~0~1~2~3
\\c)~0~2
\\d)~0~2~3
\\e)~0~2~4
\end{minipage}
}
\caption{Exercise on Reading Code Operational level \cite{lister2001}}
\label{fig:example_arch}
\end{figure}

\subsection[sec:example_layered]{An Example on Layering}\label{sec:example_layered}
An example may illustrate some of the ideas and approaches we are aiming at. Consider the exercise shown in Figure~\ref{fig:example_arch}. It comes from \cite{lister2001}. It tests the semantics of \texttt{for}-loops, the termination criteria and the operator \texttt{+=}. 

Let us get situated in the authoring role. Evidently, the exercise can be generalized in different ways by using parameterized templates such that the values of the \texttt{for} statement or the operators can be changed. In such a case, we would need a metaplan.  In order to keep the example simple we just write parts of it. We use an abstract syntax for more clarity. We consider the structure of the exercise and also the planning function in this example.%
\par Aiming at the application of genericity over the exercise structure, we may want to have a  parameterized code fragment by defining a code generating rule like the following:
\begin{center}
\begin{minipage}[hbt]{10cm}
\small
\begin{verbatim}
#genBody(init, test, limit, assign, step) =>
     for(int i=$init;i $test $limit;i $assign $step) 
        System.out.print(i+" ");
#end
\end{verbatim}
\end{minipage}
\end{center}
Thus, case c), for instance, can be generated as follows, where a distraction rule (\texttt{buggy\_limit}) is used.
\begin{center}
\begin{minipage}[hbt]{10cm}
\small
\begin{verbatim}
#caseC(init, test, limit, assign, step) =>
      c) {genBody(init, test, buggy_limit(limit), assign, step)}
#end
\end{verbatim}
\end{minipage}
\end{center}
Such kind of rules are evaluated at the MDR level, using generators targeting RCode. We intend more than textual generation occurs at the MDR level, but we just wanted to present a simple impression of the concept.%
\par To complete the example, it would remain the matter of plan representation and manipulation. Because our main concern is actually complexity, we will develop the ideas in an independent section, for more clarity.
\subsection[sec:complexity]{On Plans and Complexity}
We briefly return to example~\ref{fig:example_arch} to motivate our analysis. With respect to the intended plan in that exercise let us suppose a simple one is expected by the author. Namely, the student will check each MCQ case, sequentially, until he/she finds the correct one. In each step the student runs the body loop and check it against the number in the proposed sequence. In our conceptual model, it would mean that the student will be first processing at the DR conceptual level and then going  to the Eval level to evaluate statements at the operational and returning again to the DR level.%

\begin{figure}[htp]
\centering
\includegraphics{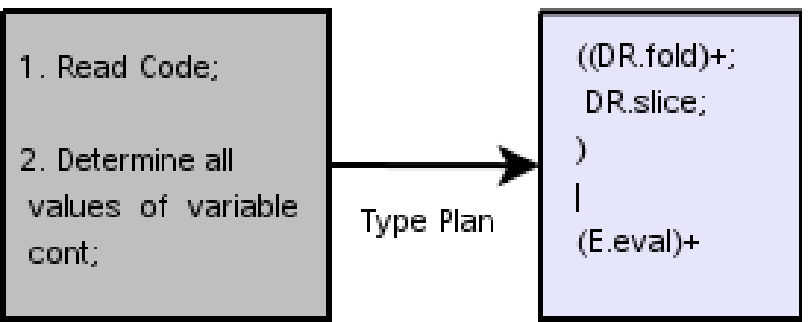}
\caption{Plan Translation and Typing}\label{fig:typeplan}
\end{figure}
\par We could certainly try to formalize such a kind of plan in a particular formal language. However, we realize that would require an independent  part of the project we are trying to define; we are not in the position of going to that level of detail, yet. Nevertheless, we think it is possible.%
\par However, we actually want to accomplish that task in a way that allows us to translate plan statements into a simpler logic language, for reasoning about complexity and eventual exercise discrepancies. Mapping the complexity of a question of an exercise to the complexity of the corresponding plan is consistent with our assumption that the complexity of an exercise is the one is expressed in its plan by its author. The idea is depicted in Figure~\ref{fig:typeplan} where some hypothetical operations on layers are assumed (named \texttt{fold}, \texttt{slice} and \texttt{eval} with the intended meanings).%
\par We figure the translation as being a kind of \textit{typing process} where the atomic types of the corresponding target formal language are \texttt{verbs} associated with cognitive layers (Eval, DR, or MDR) in our conceptual model. Such verbs will have assigned  a cognitive complexity, as in Bloom, and we will need to define two things:
\begin{itemize}
 \item How combinations of patterns denote more or less complexity, logically speaking.
\item  How complexity might affect student decisions during exercise playing.
\end{itemize}
We will illustrate and explain in more detail the idea of complexity pattern below (the concept is depicted in Figure~\ref{fig:patterns}).%
\par Because of the inherent procedural nature of plans the logic behind the typing system should be accordingly capable of handling notions like composition,  iteration and choice and the like. Hence, a dynamic-like logic seems to be a reasonable choice(\cite{harel1984,spalazzi1999}).%
\par In such a sense, we might expect that an expression denoting a plan can be of the form $P_1 | P_2$ for choice, $P_1;P_2$ for composition or $(P)^*$ for iteration, among others. As already indicated the atomic propositions would denote primitive operations on the cognitive layers. Plans can also be denoted by rules; thus $P\Rightarrow P_1 | P_2$ would be a plan that decomposes into two subplans $P_1$ and $P_2$, any of which can be followed to solve $P$, independently.
\par Just as a matter of illustration of the intended concept, we may want to express the possibility that a student could miss a part $P_2$ of a composed plan of the form $P_1;P_2$ if $P_2$ is too much smaller than the other one$P_1$, in terms of complexity. As a result, $P_2$ could  wrongly be ignored. Even though the student is able to follow each part of the plan he could fail to accomplish as a whole. We express this as follows:
\begin{mathpar}
\inferrule*[right=MissingPath($P_2$)]{\Box_{Exercise}(\Box_{Student}(P \Rightarrow P_1 ; P_2)) \\%
                                    P_1 \gg P_2
                              }
                              {\Diamond_{Student} (P \Rightarrow P_1)}
\end{mathpar}
In such a case, subplan $P_2$ is practically functioning like a strong \textit{distractor} eventually leading to a mistake. Once again, the question is whether such an \textit{eventuality}  is casual or it is so designed at purpose.  In the former, case, it is important to detect it. In the later case, how can be the exercise \textit{animated} in an \elearning environment such that the student can learn to recognize such distractors more carefully during ulterior testing stages.%
\subsection[sec:patterns_complexity]{Pattern Complexity and Reasoning Operations}
As an exercise, we want to offer a possible \textit{understanding} to events like distractors in terms of our plan complexity idea. For such a purpose, and just as an hypothesis, we assume that a \textit{rational} student will follow a plan always trying to stay at those cognitive layers which are best known to him given his knowledge and skills.
\par However, as the plan runs, some derived subtasks can force the student to change the preferred layer; in our model, for instance, if a student is at \texttt{DR} layer, he might get forced going through a \texttt{DR;E} transition or a \texttt{DR;MDR}, depending on the student \textit{cognitive preferences}. The expression \texttt{DR;E} means  to apply the operational semantics of the language while \texttt{DR;MDR} would mean to understand, in terms of some  conceptual model of reference, what the code fragment intentionally denotes.%
\par
\begin{figure}[htp]
\centering
\includegraphics{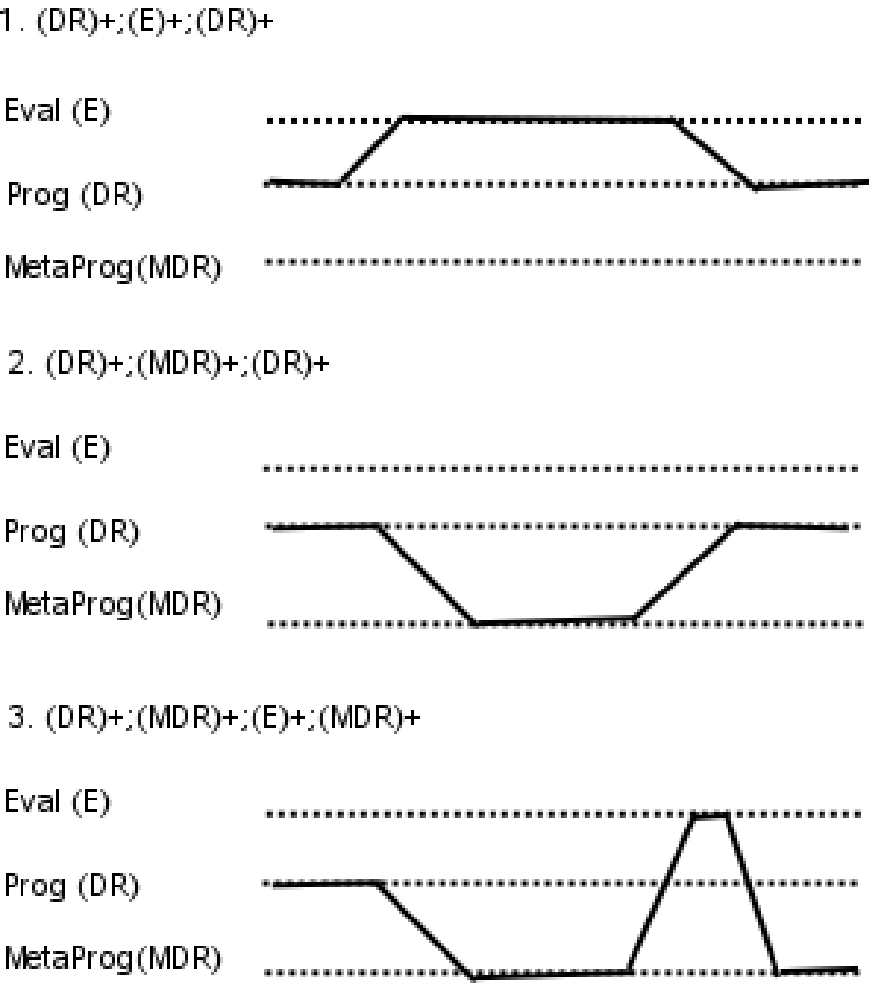}
\caption{Patterns of Complexity}\label{fig:patterns}
\end{figure}

 We illustrate some of this kind of pattern in Figure~\ref{fig:patterns}. 
We claim, for instance, that each selection of the MCQ in example in Figure~\ref{fig:example_arch} corresponds to case 1). Pattern 2) might be the solution approach an expert student would prefer to follow for the same MCQ example. Because experts (just as an assumption) could rather prefer to solve (reading-and-understanding) programming questions at more abstract levels than just mechanically computing.%
\par Moreover, we also claim that example in Figure~\ref{fig:question2} (and accompanying Figure~\ref{fig:plan_question2})  presented  at the introductory section would match pattern 3) in Figure~\ref{fig:patterns} for an expert student. We see that a transition fragment of the form \texttt{MDR;DR;E} is involved. This could mean a strong inflection; hence, it could eventually be ignored from the current plan, leading to a mistake.%
\par At the contrary, a less experienced programmer could more naturally be taking a pattern like 1), which can result in a more smooth plan, if the student understands very well the operational semantics of the language.
\vfill\pagebreak

\section[sec:conclusions]{Conclusions and Future Work}\label{sec:conclusions}
Our main objective in this work was to identify requirements and research areas for planned research  project. In retrospective, we think we have accomplished our goals in different forms and at different levels. We have developed an architectural proposal that can be useful from different perspectives, both practical and theoretically perspectives. From a particular point of view, we have gained a much better understanding of the basic elements which are the important building blocks for programming learning material into an \elearning environment like \activemath (we specifically mean programming exercises). We identified the need for a formal reasoning framework as a fundamental issue. We justified the ideas on standard cognitive models and frameworks. We think our proposal in this direction is compatible with the design principles of \activemathnc. We have identified the role that cognitive reasoning on programming can play in this context.%
\par\vspace{1.5\topsep}\noindent
Summarizing, we believe that the following specific points are important to guide future work:
\begin{itemize}
\item A review of non-standard reasoning forms not covered by Bloom's Taxonomy in terms of our proposed model.
 \item A formal language for expressing programming plans and exercises taking into consideration generative capabilities.
\item A typing system for translating plans into adequate forms of dynamic logic-like languages.
\item The precise identification of the primitive verbs and corresponding Bloom-complexity mappings (i.e. Bloom's semantics). This has to be accomplished in a cognitively sound way.
\item From a more practical perspective, the design of an extension of the \activemath OmDoc representation in order to be able to represent code at all the three layers (\texttt{Eval}, \texttt{DR}, \texttt{MDR}).
\end{itemize}

\vfill

\section*{Acknowledgments}
We gratefully thank the DAAD for the financial support that made possible 
this visit. 
My very special thanks go to \siekmannname\
for having invited me, for the contacts and the material support. 
Very special thanks to ArtinSoft 
for the support and space to make  this visit possible. 
To George Goguadze for the introduction and pointers to \activemathnc. 
And last but no least, 
I am particularly grateful to 
\autexiername,
\dietrichname, 
Marvin Schiller,  
Ewaryst Schulz and 
\wagnername\
for all the great help and the nice and friendly company.%
\vfill\pagebreak
%Bibliography
\nocite{*}
\bibliography{body}

\begin{thebibliography}{10}

\bibitem{acmieee2001}
ACM-IEEE.
\newblock {\em Computing {C}urricula 2001. {C}omputer {S}cience {V}olumen}.
\newblock ACM and IEEE, \url{http://www.sigcse.org/cc2001/}, Dec. 2001.

\bibitem{anderson2001}
L.~W. Anderson and D.~R. Krathwohl.
\newblock {\em A {T}axonomy for {L}earning, {T}eaching and {A}ssesing: {A}
  {R}evision of {Bloom}'s {T}axonomy of {E}ducational {O}bjectives}.
\newblock Longman, 2001.

\bibitem{barnes2003}
D.~Barnes and M.~K\"olling.
\newblock {\em Objects First with Java. A Practical Introduction using BlueJ}.
\newblock Prentice-Hall, 3rd edition, 2006.

\bibitem{bergin2001}
J.~Bergin.
\newblock A {P}attern {L}anguage for {I}nitial {C}ourse {D}esign.
\newblock {\em SIGCSE Bull.}, 33(1):282--286, 2001.

\bibitem{biggerstaff1993}
T.~J. Biggerstaff, B.~G. Mitbander, and D.~E. Webster.
\newblock The {C}oncept {A}ssignment {P}roblem in {P}rogram {U}nderstanding.
\newblock In {\em ICSE}, pages 482--498, 1993.

\bibitem{bordini2005}
R.~H. Bordini and J.~F. H{\"u}bner.
\newblock {BDI} {A}gent {P}rogramming in {AgentSpeak} using {Jason} ({T}utorial
  {P}aper).
\newblock In Toni and Torroni \cite{DBLP:conf/clima/2005}, pages 143--164.

\bibitem{brachman2004}
R.~J. Brachman and H.~J. Levesque.
\newblock {\em Knowledge Representation and Reasoning}.
\newblock Elsevier, 2004.

\bibitem{buck2000}
D.~Buck and D.~J. Stucki.
\newblock Design {E}arly {C}onsidered {H}armful: graduated exposure to
  complexity and structure based on levels of cognitive development.
\newblock {\em SIGCSE Bull.}, 32(1):75--79, 2000.

\bibitem{distefano2004}
A.~{Di\, Stefano} and C.~Santoro.
\newblock {D}esigning {C}ollaborative {A}gents with e{XAT}.
\newblock In {\em Second International Workshop on Agent-based Computing for
  Enterprise Collaboration}, June 2004.

\bibitem{emir2007}
B.~Emir, M.~Odersky, and J.~Williams.
\newblock Matching {O}bjects with {P}atterns.
\newblock In {\em ECOOP 2007}, volume 4609 of {\em LNCS}. Springer, 2007.

\bibitem{felleisen2003}
M.~Felleisen, R.~B. Findler, M.~Flatt, and S.~Krishnamurthi.
\newblock {The TeachScheme! Project: Computing and Programming for Every
  Student}.

\bibitem{felleisen1998}
M.~Felleisen, R.~B. Findler, M.~Flatt, and S.~Kristnamurthi.
\newblock The {DrScheme} {P}roject: {A}n {O}verview.
\newblock {\em SIGPLAN Notices}, 33(6):17--23, 1998.

\bibitem{labra2003}
J.~E.~L. Gayo, J.~M.~M. Gil, A.~M.~F. \'{A}lvarez, and H.~S. Chigne.
\newblock A {G}eneric {E}-learning {M}ultiparadigm {P}rogramming {L}anguage
  {S}ystem: {IDEFIX} {P}roject.
\newblock {\em SIGCSE Bull.}, 35(1):391--395, 2003.

\bibitem{germain2006}
G.~Germain.
\newblock Concurrency {O}riented {P}rogramming in {Termite} {Scheme}.
\newblock In {\em 2006 ACM SIGPLAN Workshop on Erlang}, pages 20--20. ACM,
  2006.

\bibitem{glaser2000}
H.~Glaser, P.~H. Hartel, and P.~W. Garratt.
\newblock Programming by {N}umbers: A {P}rogramming {M}ethod for {N}ovices.
\newblock {\em Comput. J.}, 43(4):252--265, 2000.

\bibitem{icce05}
G.~Goguadze, A.~G. Palomo, and E.~Melis.
\newblock {ActiveMath}.
\newblock In {\em International Conference on Computers in Education (ICCE
  2005)}, Singapore, 2005.

\bibitem{gouadze2007}
G.~Goguadze and J.~Tsigler.
\newblock Authoring {I}nteractive {E}xercises in {ActiveMath}.
\newblock In {\em MathUI Workshop at Mathematical Knowledge Management},
  \url{http://www.examat.org/}, June 2007.

\bibitem{gonzales2006}
G.~Gonzalez.
\newblock A {S}ystematic {A}pproach to {A}ctive and {C}ooperative {L}earning in
  {CS1} and its {E}ffects on {CS2}.
\newblock {\em SIGCSE Bull.}, pages 133--137, 2006.

\bibitem{DBLP:conf/ace/2003}
T.~Greening and R.~Lister, editors.
\newblock {\em Fifth Australasian Computing Education Conference (ACE 2003),
  Adelaide, Australia, 4-7 February 2003}, volume~20 of {\em CRPIT}. Australian
  Computer Society, 2003.

\bibitem{DBLP:conf/sigcse/2003}
S.~Grissom, D.~Knox, D.~Joyce, and W.~Dann, editors.
\newblock {\em Proceedings of the 34th SIGCSE Technical Symposium on Computer
  Science Education, 2003, Reno, Nevada, USA, February 19-23, 2003}. ACM, 2003.

\bibitem{guzdial2002}
M.~Guzdial and E.~Soloway.
\newblock {T}eaching the {Nintendo} {G}eneration to {P}rogram.
\newblock {\em Commun. ACM}, 45(4):17--21, 2002.

\bibitem{harel1984}
D.~Harel.
\newblock {Dynamic Logic}.
\newblock In D.~Gabbay and F.~Guenther, editors, {\em Handbook of Philosophical
  Logic Volume {II} --- Extensions of Classical Logic}, pages 497--604.
  D.~Reidel Publishing Company: Dordrecht, The Netherlands, 1984.

\bibitem{hocgreen1991}
J.-M. Hoc, G.~T.~R. G., D.~Gilmore, and R.~Samurçay, editors.
\newblock {\em The {P}sychology of {P}rogramming}.
\newblock Academic Press, London, 1991.

\bibitem{holohan2006}
E.~Holohan, M.~Melia, D.~McMullen, and C.~Pahl.
\newblock The {G}eneration of {E}-{L}earning {E}xercise {P}roblems from
  {S}ubject {O}ntologies.
\newblock In {\em Sixth Conference International Conference on Advanced
  Learning Technologies}, 2006.

\bibitem{hundhausen2006}
C.~D. Hundhausen, J.~L. Brown, S.~Farley, and D.~Skarpas.
\newblock A {M}ethodology for {A}nalyzing the {T}emporal {E}volution of
  {N}ovice {P}rograms {B}ased on {S}emantic {C}omponents.
\newblock In {\em ICER '06: Proceedings of the 2006 international workshop on
  Computing education research}, pages 59--71, New York, NY, USA, 2006. ACM.

\bibitem{jenkins2001}
T.~Jenkins.
\newblock The {M}otivation of {S}tudents of {P}rogramming.
\newblock {\em SIGCSE Bull.}, 33(3):53--56, 2001.

\bibitem{johnson2007}
C.~G. Johnson and U.~Fuller.
\newblock Is {B}loom's {T}axonomy {A}ppropriate for {C}omputer science?
\newblock In A.~Berglund and M.~Wiggberg, editors, {\em Proceedings of the
  Sixth Baltic Sea Conference on Computing Education Research}, volume 2007-006
  of {\em Uppsala University Department of Information Technology Technical
  Reports}, pages 120--123. Uppsala University, February 2007.

\bibitem{kohlhasecodeml2008}
M.~Kohlhase.
\newblock {\em CodeML: {A}n {O}pen {M}arkup {F}ormat the {C}ontent and
  {P}resentatation of {P}rogram Code; the {CodeML} {S}pecification.}
\newblock OmDoc.org,
  \url{https://svn.omdoc.org/repos/codeml/doc/spec/codeml.pdf}, 2008.

\bibitem{kohlhase2008}
M.~Kohlhase.
\newblock {\em An {O}pen {M}arkup {F}ormat for {M}athematical {D}ocuments,
  {V}ersion {1.2}}.
\newblock Number 4180 in LNAI. Springer,
  \url{http://www.omdoc.org/pubs/omdoc1.2.pdf}, April 2008.

\bibitem{kuittinen2004}
M.~Kuittinen and J.~Sajaniemi.
\newblock Teaching {R}oles of {V}ariables in {E}lementary {P}rogramming
  {C}ourses.
\newblock In {\em ITiCSE '04: Proceedings of the 9th annual SIGCSE conference
  on Innovation and technology in computer science education}, pages 57--61,
  New York, NY, USA, 2004. ACM.

\bibitem{lister2001}
R.~Lister.
\newblock Objectives and {O}bjective {A}ssesment in {CS1}.
\newblock In {\em SIGCSE}, pages 292--296, 2001.

\bibitem{lister2003}
R.~Lister and J.~Leaney.
\newblock {F}irst {Y}ear {P}rogramming: {L}et {A}ll the {F}lowers {Bloom}.
\newblock In Greening and Lister \cite{DBLP:conf/ace/2003}, pages 221--230.

\bibitem{lister2003a}
R.~Lister and J.~Leaney.
\newblock Introductory {P}rogramming, {C}riterion-referencing, and {Bloom}.
\newblock In Grissom et~al. \cite{DBLP:conf/sigcse/2003}, pages 143--147.

\bibitem{lister2006SOLO}
R.~Lister, B.~Simon, E.~Thompson, J.~L. Whalley, and C.~Prasad.
\newblock Not seeing the forest for the trees: novice programmers and the solo
  taxonomy.
\newblock {\em SIGCSE Bull.}, 38(3):118--122, 2006.

\bibitem{mccracken2001}
M.~McCracken, V.~Almstrum, D.~Diaz, M.~Guzdial, D.~Hagan, Y.~B.-D. Kolikant,
  C.~Laxer, L.~Thomas, I.~Utting, and T.~Wilusz.
\newblock A {M}ulti-national, {M}ulti-institutional {S}tudy of {A}ssessment of
  {P}rogramming {S}kills of {F}irst-year {CS} {S}tudents.
\newblock In {\em ITiCSE-WGR '01: Working group reports from ITiCSE on
  Innovation and technology in computer science education}, pages 125--180, New
  York, NY, USA, 2001. ACM.

\bibitem{Melisetal-ActiveMath-AIEDJ-2001}
E.~Melis, E.~Andr\'es, J.~B\"udenbender, A.~Frischauf, G.~Goguadze,
  P.~Libbrecht, M.~Pollet, and C.~Ullrich.
\newblock Activemath: {A} {G}eneric and {A}daptive {W}eb-{B}ased {L}earning
  {E}nvironment.
\newblock {\em International Journal of Artificial Intelligence in Education},
  12(4):385--407, 2001.

\bibitem{meyer2006}
B.~Meyer.
\newblock {T}estable, {R}eusable {U}nits of {C}ognition.
\newblock {\em IEEE Computer}, 39(4):20--24, 2006.

\bibitem{pedroni2006}
M.~Pedroni and B.~Meyer.
\newblock The {I}nverted {C}urriculum in {P}ractice.
\newblock {\em SIGCSE Bull.}, 38(1):481--485, 2006.

\bibitem{powers2006}
K.~Powers, P.~Gross, S.~Cooper, M.~McNally, K.~J. Goldman, V.~Proulx, and
  M.~Carlisle.
\newblock Tools for {T}eaching {I}ntroductory {P}rogramming: {W}hat works?
\newblock {\em SIGCSE Bull.}, 38(1):560--561, 2006.

\bibitem{robinet2007}
V.~Robinet, G.~Bisson, M.~Gordon, and B.~Lemaire.
\newblock Searching for {S}tudent {I}ntermediate {S}teps.
\newblock In {\em UM2007: Workshop On Data Mining For User Modeling}, June
  2007.

\bibitem{kumar2006}
C.~K. Roy, T.~Noll, B.~Roy, and J.~R. Cordy.
\newblock {T}owards {A}utomatic {V}erification of {Erlang} {P}rograms by
  {$\pi$-Calculus} {T}ranslation.
\newblock In {\em 2006 ACM SIGPLAN Workshop on Erlang}, pages 38--50. ACM,
  2006.

\bibitem{smolka2008}
G.~Smolka.
\newblock {\em {Programmierung: Eine Einf\"uhrung in die Informatik mit
  Standard ML}}.
\newblock Odenbourg.de, 2008.

\bibitem{solloway1986}
E.~Soloway.
\newblock Learning to {P}rogram = {L}earning to {C}onstruct {M}echanisms and
  {E}xplanations.
\newblock {\em Commun. ACM}, 29(9):850--858, 1986.

\bibitem{spalazzi1999}
L.~Spalazzi and P.~Traversi.
\newblock A {D}ynamic {L}ogic for {A}cting, {S}ensing and {P}lanning.
\newblock {\em Journal of Logic Computation}, pages 1--36, 1999.

\bibitem{thompson2008}
E.~Thompson, A.~Luxton-Reilly, J.~L. Whalley, M.~Hu, and P.~Robins.
\newblock Blomm's {T}axonomy for {CS}{A}ssesment.
\newblock In {\em Tenth Australasian Computing Education Conference ACE2008},
  2008.

\bibitem{DBLP:conf/clima/2005}
F.~Toni and P.~Torroni, editors.
\newblock {\em Computational Logic in Multi-Agent Systems, 6th International
  Workshop, CLIMA VI, London, UK, June 27-29, 2005, Revised Selected and
  Invited Papers}, volume 3900 of {\em Lecture Notes in Computer Science}.
  Springer, 2006.

\bibitem{wikipedia:elearning}
\url{http://en.wikipedia.org/wiki/Elearning}.
\newblock Electronic {L}earning.

\bibitem{wiedenbeck1999}
S.~Wiedenbeck.
\newblock Organization of {P}rogramming {K}nowledge of {N}ovices and {E}xperts.
\newblock {\em Journal of the American Society for Information Science},
  37(5):294--299, 1999.

\bibitem{winslow1996}
L.~E. Winslow.
\newblock Programming {P}edagogy {--}{A} {P}sychological {O}verview.
\newblock {\em SIGCSE Bull.}, 28(3):17--22, 1996.

\bibitem{zinn2006}
C.~Zinn.
\newblock Supporting {T}utorial {F}eedback to {S}tudent {H}elp {R}equests and
  {E}rrors in {S}ymbolic {D}ifferentiation.
\newblock In M.~Ikeda and K.~Ashley, editors, {\em Proceedings of Intelligent
  Tutoring Systems 8th. International Conference ITS-2006}, volume LNCS 4053 of
  {\em Lecture Notes in Computer Science}, pages 349--359. Springer-Verlag,
  June 2006.

\end{thebibliography}
\bibliographystyle{abbrv}
\end{document}